\documentclass[10pt,a4paper]{article}

\usepackage{nxtlab-tech-report}

\reporttitle{MRMS: A Multi-Resolution Memory Substrate for Long-Lived AI Agents}
\reportauthor{Jizhizi Li \quad Amy Shi-Nash}
\reportteam{NxtLab Innovations}
\reportdate{July 2026}
\reportstatus{Technical Report}
\reportcontact{support@nxtlab.com.au}
\reportkeywords{long-lived agents, lifelong agents, agent memory, memory-augmented language models, structured-vector-graph memory, memory consolidation, context projection, personalization, retrieval, knowledge graphs}

\reportabstract{
Long-lived AI agents require continuity across interactions, but continuity cannot be obtained by simply extending the prompt window.
An agent must preserve useful prior experience, retrieve it selectively, distinguish personal context from external evidence, and revise memory when the underlying situation changes.
We propose an architectural memory substrate organized along two orthogonal axes: a representational axis spanning structured records, vector representations, and graph relations; and a temporal axis spanning short-term traces, medium-term abstractions, and long-term semantic commitments.
Its key design constraint is synchronized structured-vector-graph memory: structured records govern eligibility, vector representations support recall, and graph relations adjudicate support, contradiction, and supersession before gated context projection.
Its central claim is that reliable personalization is a memory design problem: useful memory is structured, selectively exposed, continuously consolidated, and epistemically labeled rather than stored as undifferentiated conversation history.
Beyond the framework, we instantiate MRMS as a lightweight prototype implementing structured records, vector retrieval, temporal policies, and graph-based revision.
The prototype exercises the core substrate mechanisms through pre-generation memory selection, revision, boundary enforcement, and evidence attribution under controlled long-lived interaction scenarios with explicit evidence requirements.
}

\begin{document}

\makenxtlabpapertitle

\section{Introduction}

Language models can produce coherent responses within a single prompt, yet long-lived agents require a different form of competence.
They must preserve goals, preferences, constraints, prior decisions, and unresolved tasks across a sequence of interactions.
This requirement creates a stability-plasticity tension: an agent should be plastic enough to incorporate new evidence, but stable enough not to rewrite its representation of a user or task after every turn.
It should remember what matters, forget what is incidental, and expose only the memory that is appropriate for the present situation.

The simplest approach is to enlarge the context window and place more history into the prompt.
This is useful but incomplete.
Long contexts remain expensive, attention over long text can be uneven, and irrelevant history can distract the model from the current task \citep{liu2024lost}.
More importantly, a transcript is not a memory system.
It does not distinguish a durable preference from a temporary instruction, a verified fact from a speculative inference, or a private observation from external evidence.

These distinctions matter because memory errors in long-lived agents are persistent rather than local.
A false negative loses continuity, but a false positive can be more damaging: stale preferences, superseded facts, or out-of-scope observations may be repeatedly reintroduced into future reasoning.
Reliable memory therefore requires governing influence, not simply maximizing recall.
Before a memory object reaches the prompt, the system should know where it came from, whether it is still active, which boundary it belongs to, and whether newer evidence supports, revises, or supersedes it.
This shifts agent memory from passive storage to pre-generation control.
The central question is no longer only whether a relevant item can be found, but whether the item should be allowed to influence the next action, how its provenance should be exposed, and how it should interact with newer or conflicting evidence.

We frame continuity as a memory substrate problem.
A memory substrate is the representational and operational layer that stores interaction traces, compresses them into reusable abstractions, retrieves them under constraints, and revises them as evidence changes.
The substrate is not a monolithic database and not a hidden prompt.
It is a set of memory objects with different temporal scales, source classes, confidence levels, and update rules.

\paragraph{Contributions.}
This paper is best read as an architectural position paper with a diagnostic benchmark.
Our contributions are fourfold: (1) we introduce MRMS, a two-axis memory substrate that jointly organizes agent memory along representational and temporal dimensions; (2) we propose a synchronized structured-vector-graph representation that separates semantic retrieval from authorization and relational reasoning; (3) we define synchronization invariants and boundary-aware context projection for reliable long-lived personalization; and (4) we instantiate the framework as a prototype with a diagnostic probe evaluating revision, stale suppression, attribution, and boundary control.

\begin{figure}[t]
  \centering
  \includegraphics[width=\linewidth]{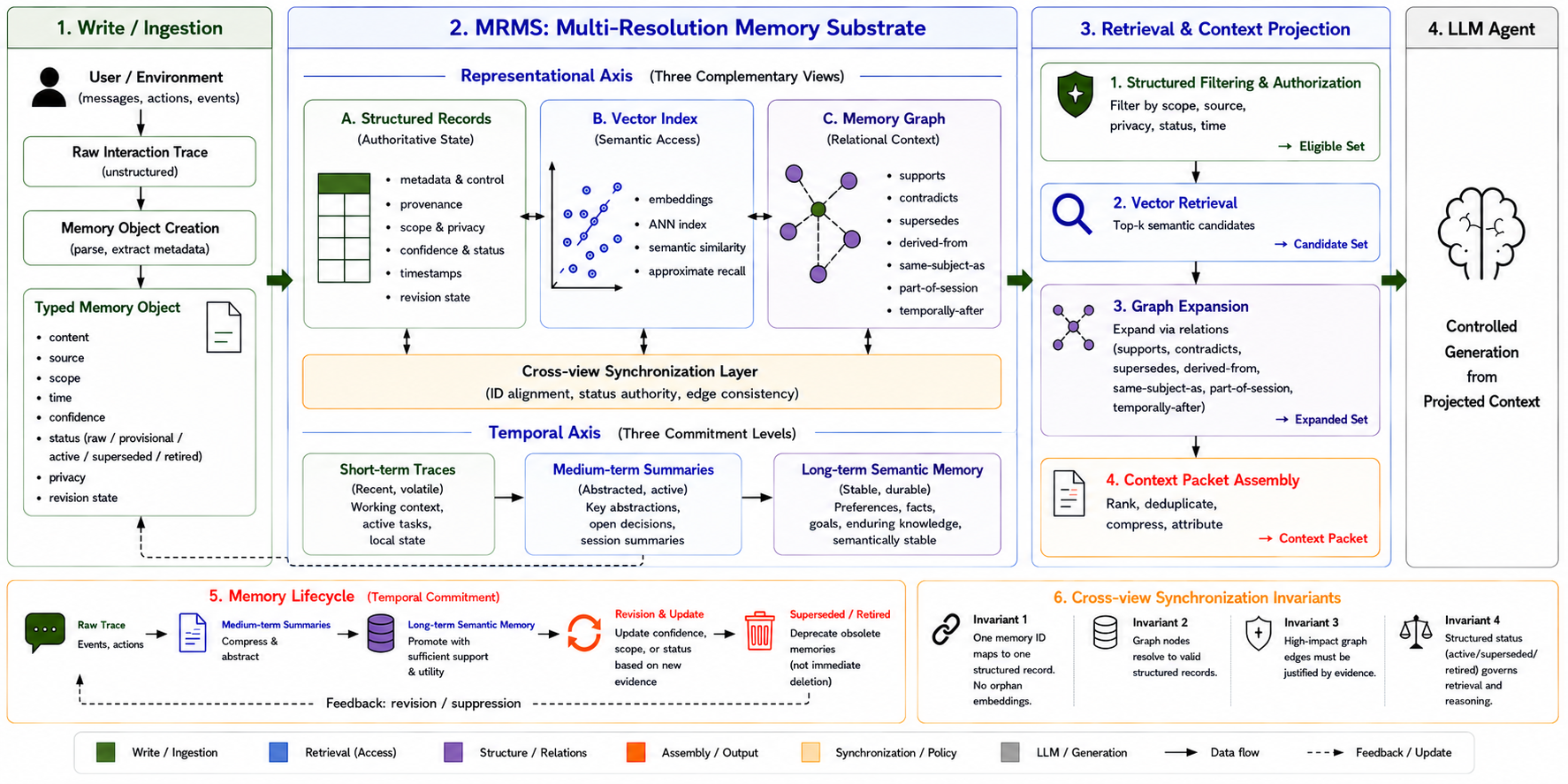}
  \caption{MRMS architecture: write/ingestion, synchronized memory views, retrieval, lifecycle, and cross-view invariants.}
  \label{fig:mrms-architecture}
\end{figure}

\section{Related Work}

\paragraph{Cognitive memory as design analogy.}
The distinction between short-term, long-term, episodic, and semantic memory predates modern AI systems.
Atkinson and Shiffrin proposed a multi-store view of memory in which information moves through sensory registers, short-term storage, and long-term storage under control processes \citep{atkinson1968human}; Baddeley and Hitch characterized working memory as active maintenance for ongoing cognition \citep{baddeley1974working}.
Tulving distinguished episodic memory, concerning situated events, from semantic memory, concerning general knowledge and stable meaning \citep{tulving1972episodic}.
These theories are not adopted here as biological claims; they highlight design pressures for artificial agents: local context should be immediately available, episodic traces should preserve time and source, and semantic memory should support abstraction.

The complementary learning systems hypothesis is also relevant.
McClelland, McNaughton, and O'Reilly argued that cognition benefits from fast-learning and slow-learning systems, where one preserves recent experience and the other gradually forms stable structure \citep{mcclelland1995complementary}.
For AI memory, this suggests that raw traces, summaries, and semantic facts should have different update frequencies.
Immediate memory should be easy to write and discard; long-term memory should require stronger evidence, support revision, and carry provenance.
Earlier neural memory architectures also studied explicit read-write memory \citep{weston2015memory,sukhbaatar2015end}, but MRMS treats lifecycle state, source boundaries, and graph relations as explicit interfaces rather than learned state.

\paragraph{Retrieval, long context, and agent memory.}
Retrieval-augmented generation separates parametric knowledge from external evidence.
Retrieval-augmented models combine dense retrieval with generation and show that models need not internalize all knowledge in parameters \citep{guu2020realm,lewis2020retrieval,karpukhin2020dense}, while approximate nearest-neighbor indexes make large retrieval stores practical \citep{malkov2020efficient,johnson2021billion}.
Classical retrieval is usually oriented toward documents rather than personal continuity.
Long-lived agents must retrieve prior decisions, preferences, unresolved goals, and consolidated facts, while keeping these channels distinct from external evidence.
The graph component of MRMS is closer in spirit to knowledge-graph and relational-learning views, where typed relations preserve structure that is not recoverable from vector proximity alone \citep{nickel2016review,hogan2021knowledge}.

Long-context systems also help but do not remove the need for memory management.
Transformer-XL extends attention through recurrence \citep{dai2019transformerxl}, and LongMem augments language models with a long-term memory cache \citep{wang2023longmem}.
However, longer context windows do not automatically yield better memory: relevant information can be missed or diluted in long sequences \citep{liu2024lost}.
MemGPT therefore treats prompt context as scarce and external memory as something that must be explicitly managed \citep{packer2023memgpt}.

Agentic memory adds consolidation, reflection, and persistent state.
Generative Agents stored observations and reflected on them for planning \citep{park2023generative}; Reflexion stored verbal feedback about failures \citep{shinn2023reflexion}; and related agent work externalizes state through reasoning traces, tool use, and self-feedback \citep{yao2023react,schick2023toolformer,madaan2023selfrefine}.
Personal and production-oriented memory systems such as MemoryBank, Mem0, Zep/Graphiti, A-MEM, HippoRAG, and Letta/MemGPT emphasize long-term user memory, graph organization, or stateful agent services \citep{zhong2024memorybank,chhikara2025mem0,rasmussen2025zep,xu2025amem,gutierrez2024hipporag,packer2023memgpt}.
Older cognitive architectures such as Soar and ACT-R also separate memory structures from control processes \citep{laird1987soar,anderson2004integrated}.
Conversational memory benchmarks such as LoCoMo \citep{maharana2024locomo}, LongMemEval \citep{wu2025longmemeval}, and MemoryAgentBench \citep{hu2025memoryagentbench} shift evaluation toward memory over time.
However, recall alone is insufficient: a good substrate must also decline to use memory when it is stale, weak, or outside scope.

\paragraph{What distinguishes MRMS?}
Existing memory systems already combine structured records, vector search, graph relations, and stateful services in different ways.
MRMS instead focuses on governing memory influence.
Rather than treating memory as a homogeneous collection of retrieved snippets, MRMS separates structured control state, semantic addressability, and relational validity.
This separation allows memory objects to remain retrievable while being restricted by authorization, temporal validity, provenance, and supersession policies.
The resulting substrate targets reliable long-lived behavior rather than retrieval quality alone.
MRMS is therefore complementary to RAG, long-context models, and agent reflection: those mechanisms supply candidate context, while MRMS governs whether and how memory may influence future behavior.
It is also complementary to continual parameter adaptation: MRMS externalizes evolving state into auditable memory objects, allowing an agent to adapt its context without silently rewriting the underlying model.
We do not frame the diagnostic benchmark as a direct head-to-head comparison with deployed systems such as MemGPT, MemoryBank, Mem0, or Zep: their interfaces typically expose retrieved text or final behavior, not pre-generation memory-object decisions with status, scope, supersession, and attribution labels.
We therefore evaluate controlled MRMS ablations to isolate substrate mechanisms.

\section{Problem Setting}

Let an interaction at turn $t$ be represented as
\[
e_t = (u, x_t, y_t, \tau_t, \pi_t),
\]
where $u$ denotes the interacting subject, $x_t$ is the input, $y_t$ is the agent response, $\tau_t$ is time, and $\pi_t$ is the local policy or task context.
A long-lived agent maintains a memory state
\[
M_t = (W_t, S_t, E_t, G_t, K_t),
\]
where $W_t$ is working memory, $S_t$ is session memory, $E_t$ is episodic memory, $G_t$ is semantic memory, and $K_t$ is external knowledge.

The generation objective is not to condition on all of $M_t$.
It is to construct a small evidence set
\[
C_t = \operatorname{select}(x_t, M_t, \pi_t)
\]
such that $C_t$ is relevant to the present input, consistent with applicable boundaries, and concise enough to support reliable reasoning.
The response can then be viewed as
\[
y_t \sim p_\theta(\cdot \mid x_t, C_t, \pi_t).
\]
After generation, memory is updated by a consolidation operator:
\[
M_{t+1} = \operatorname{consolidate}(M_t, e_t).
\]
Selection determines what the model should see now; consolidation determines what the system should preserve for later.
A memory object may be stored without being selected, and an object may be selected without being permanently reinforced.

\section{Two-Axis Memory Substrate}

We propose a two-axis, multi-resolution memory substrate (MRMS).
The method is intended as an architectural abstraction rather than a single concrete realization.
Its central idea is to organize memory through two orthogonal axes.
The \emph{representational axis} determines how a memory is encoded and made usable: as a structured object, a vector representation, and a graph relation.
The \emph{temporal axis} determines how strongly and persistently a memory should influence future behavior: from short-term interaction traces, to medium-term session abstractions, to long-term semantic commitments.
Figure~\ref{fig:mrms-architecture} summarizes how these axes interact with selection, context projection, consolidation, and revision.

\subsection{Memory objects and structured state}

Each memory object is represented as
\[
m_i = (\ell_i, q_i, v_i, r_i, a_i, \sigma_i, c_i, h_i),
\]
where $\ell_i$ is the memory layer, $q_i$ is a compact textual claim or summary, $v_i$ is a retrieval representation, $r_i$ records source and time, $a_i$ stores attribution and provenance, $\sigma_i$ defines scope, $c_i$ is confidence, and $h_i$ is revision history.
The textual claim makes the memory inspectable, the vector representation makes it retrievable, the source record makes it accountable, and the revision history prevents updates from becoming silent overwrites.

The structured part of the representation is not merely a container for text.
It is the control plane of the memory system.
It stores the current status of the object, its temporal layer, the subject and task scope under which it may be used, its source class, and a revision log.
These fields determine whether the object is eligible for retrieval, promotion, suppression, or audit.
In an implementation, this role naturally corresponds to a document or record store: records can be updated atomically, queried by metadata, and retained even when their vector or graph views are rebuilt.
The theoretical requirement is not a specific database, but the existence of an inspectable structured record that remains the authoritative state of the memory object.

A memory object may have several statuses.
\emph{Raw} memories preserve interaction evidence with minimal interpretation.
\emph{Provisional} memories express inferred patterns that require additional support.
\emph{Active} memories may be used for generation when selected under the current boundary.
\emph{Superseded} memories are retained as historical context but should not guide current behavior.
\emph{Retired} memories are suppressed unless explicitly needed for audit or historical reconstruction.
Revision is therefore not always deletion: an outdated preference can be false as a current guide while still being true as a record of what was once expressed.

\subsection{Structured-vector-graph memory}

The distinctive representational feature of MRMS is that memory is indexed in three complementary ways.
The structured store $\mathcal{S}_t$ maintains authoritative records:
\[
\mathcal{S}_t = \{(i, \ell_i, q_i, r_i, a_i, \sigma_i, c_i, h_i, \mathrm{status}_i): m_i \in M_t\}.
\]
This store supports exact filters over subject, source, scope, status, and temporal layer.
It answers questions such as whether a memory is active, who or what it is about, whether it came from interaction or external evidence, and whether it has been superseded.
These filters are intentionally applied before generation because semantic similarity alone cannot determine authorization.

A vector index then supports approximate semantic recall:
\[
\mathcal{V}_t = \{(i, v_i): m_i \in M_t\}.
\]
This index is useful when the present input resembles an earlier situation without sharing exact words.
However, vector similarity alone is not a sufficient theory of memory.
Two memories can be close in embedding space while differing in scope, truth status, source, or temporal validity.

MRMS therefore pairs the structured and vector views with a typed memory graph:
\[
\mathcal{G}_t = (V_t, A_t, \rho),
\]
where each node in $V_t$ is a memory object and each edge in $A_t$ has a relation type assigned by $\rho$.
Useful relation types include \emph{supports}, \emph{contradicts}, \emph{supersedes}, \emph{derived-from}, \emph{same-subject-as}, \emph{part-of-session}, and \emph{temporally-after}.
The graph expresses relational validity: not just whether a memory is relevant, but how it is connected to other claims, traces, corrections, and evidence.

\begin{table}[t]
  \centering
  \small
  \begin{tabularx}{\linewidth}{@{}p{0.22\linewidth}p{0.27\linewidth}X@{}}
    \toprule
    View & Primary query & Failure prevented \\
    \midrule
    Structured record & Exact filters over source, scope, status, time, layer & Unauthorized retrieval, silent overwrite, missing provenance. \\
    Vector index & Approximate semantic neighborhood search & Lexical mismatch and brittle exact lookup. \\
    Memory graph & Typed traversal over support, contradiction, supersession, and session edges & Stale recall, unresolved conflict, overgeneralization. \\
    \bottomrule
  \end{tabularx}
  \caption{Complementary roles of structured, vector, and graph memory views.}
  \label{tab:svg-views}
\end{table}

Given an input $x_t$, selection first applies structured gates to obtain an authorized set:
\[
\mathcal{A}_t = \{m_i \in M_t : \operatorname{auth}(m_i,u,\pi_t)=1,\ \operatorname{active}(m_i)=1\}.
\]
Vector search then produces semantic candidates within that authorized set:
\[
\mathcal{N}_k(x_t)=\operatorname{topk}_{m_i \in \mathcal{A}_t}\operatorname{sim}(f(x_t),v_i).
\]
The graph then expands and filters this set by following typed edges to supporting traces, contradictions, superseding updates, and scope constraints.
The generator receives a structured context packet containing the selected claim, evidence, active scope, and known conflict, rather than an unordered list of snippets.

\subsection{Temporal commitment}

The second axis controls the degree of temporal commitment.
Short-term traces are cheap to write and easy to discard.
Medium-term summaries compress bounded interaction into a reusable state.
Long-term commitments influence future behavior and therefore require stronger evidence.
This distinction is important because the same representational form can appear at different temporal scales.
A short-term memory may be a structured trace with a vector descriptor but no durable graph edges.
A medium-term memory may be a session summary linked to open decisions.
A long-term memory may be a semantic claim supported by several traces and guarded by contradiction edges.

Promotion across temporal layers should therefore be conservative.
MRMS allows rapid capture in lower layers while requiring stronger support for durable influence.
This is the main stability mechanism in the framework: the system can remember quickly without immediately believing permanently.
The substrate can record that something was said, later summarize why it mattered, and only then promote it into a semantic commitment if repeated evidence and task utility justify the risk.
Appendix~\ref{app:benchmark-details} gives an implementation-facing breakdown of these temporal layers across the structured, vector, and graph views.

\subsection{Cross-view synchronization}

The three representational views must remain synchronized without collapsing into a single store.
MRMS assigns each memory object a stable identifier $i$ that acts as the join key across structured, vector, and graph views.
The structured record is authoritative for lifecycle state; the vector index is authoritative only for semantic addressability; the graph is authoritative for relational interpretation.
This prevents a common failure in agent memory systems: a stale embedding remains retrievable even after the underlying claim has been corrected.
In MRMS, retrieval by vector similarity must be checked against the structured status field and graph supersession edges before the memory can influence generation.

Synchronization is governed by four invariants.
First, every vector point must resolve to exactly one structured memory record.
If the structured record is retired or superseded, the vector point may remain stored for audit or offline maintenance, but it cannot be selected as active context.
Second, every graph node that can be traversed during selection must resolve to a structured memory record.
This ensures that graph expansion cannot reintroduce unauthorized or stale claims.
Third, high-impact graph edges must be justified by evidence-bearing records.
For example, a \emph{supersedes} edge should point from a newer memory object to the older object it replaces, and the newer object should contain the source event or summary that motivated the update.
Fourth, structured status dominates retrieval and reasoning: retired or superseded records cannot guide generation even if they remain semantically retrievable or graph-adjacent.

\begin{table}[t]
  \centering
  \small
  \begin{tabularx}{\linewidth}{@{}p{0.25\linewidth}X@{}}
    \toprule
    Invariant & Purpose \\
    \midrule
    Stable cross-view identifier & Allows structured records, vector points, and graph nodes to be joined during retrieval and audit. \\
    Graph node validity & Ensures graph expansion cannot traverse memory objects that lack valid structured records. \\
    Evidence-backed relation edges & Ensures support, contradiction, and supersession edges remain explainable rather than latent labels. \\
    Structured status dominates selection & Prevents stale vector hits or graph traversals from bypassing retirement, scope, or supersession. \\
    \bottomrule
  \end{tabularx}
  \caption{Synchronization invariants for the structured-vector-graph substrate.}
  \label{tab:sync-invariants}
\end{table}

This synchronization design also gives the substrate a practical maintenance model.
Embeddings can be rebuilt after an encoder change because the structured store retains the canonical record.
Graph edges can be recomputed or pruned because the evidence records remain inspectable.
Structured records can be compacted because vector and graph views preserve retrieval and relation signals.
The memory system therefore avoids binding its theoretical definition to a particular storage engine while still imposing concrete consistency requirements.
We view these synchronization invariants as the primary conceptual contribution of MRMS.
Previous memory systems often combine multiple storage mechanisms, but typically treat them as implementation choices.
MRMS instead elevates synchronization between structured state, vector retrieval, and relational reasoning into an explicit architectural invariant that governs memory correctness over long-lived interactions.

\section{Operations}

MRMS follows a five-stage cycle.
The current interaction is parsed into a trace; candidate memories are written as low-commitment records; selection constructs a bounded context projection; consolidation transforms traces into summaries, stable propositions, relations, or procedural lessons; and revision updates existing objects when later evidence contradicts, narrows, or supersedes them.

\paragraph{Write.}
The write operation records an interaction trace,
\[
\operatorname{write}(M_t, e_t) \rightarrow M_t^+.
\]
Writing is weaker than consolidation: it records that an event occurred, but does not imply that the event should become a durable fact.
A write produces at least two artifacts, an evidence-preserving trace and a retrieval-oriented descriptor.
In the structured-vector-graph setting, a write may update all three views but with different levels of commitment.
The structured record is created first and receives a stable memory identifier.
The vector view receives an embedding of the claim, summary, or trace descriptor.
The graph view receives only low-risk edges at first, such as \emph{part-of-session} or \emph{derived-from}.
Higher-impact edges such as \emph{supersedes} and \emph{contradicts} can be added during consolidation or revision after additional checks.

\paragraph{Select.}
Selection projects memory into the generation context,
\[
C_t = \operatorname{select}(x_t, \tilde{M}_t, b_t),
\]
where $\tilde{M}_t$ is the memory available under current boundary conditions and $b_t$ denotes a budget.
Selection applies hard gates for scope, authorization, and status; ranks eligible objects by relevance, confidence, freshness, and marginal utility; traverses graph neighborhoods for evidence and conflict; and assembles a context packet that separates personal memory from external evidence.

For each eligible object, the substrate estimates
\[
s_i =
\alpha R(m_i,x_t)
+ \beta C(m_i)
+ \gamma F(m_i,t)
+ \delta U(m_i,\pi_t)
- \eta N(m_i,C_t),
\]
where $R$ is semantic relevance, $C$ is confidence, $F$ is freshness, $U$ is expected task utility, and $N$ is redundancy or distraction cost.
The selected context should be compact, high-confidence, nonredundant, and explicitly scoped.
We further decompose retrieval into four stages:
\[
x_t \rightarrow \mathcal{A}_t \rightarrow \mathcal{N}_k(x_t) \rightarrow \mathcal{H}_t \rightarrow C_t,
\]
where $\mathcal{A}_t$ is the structured eligible set, $\mathcal{N}_k$ is the vector neighborhood, $\mathcal{H}_t$ is the graph-expanded evidence subgraph, and $C_t$ is the final context packet.
This decomposition is useful for diagnosis: an error can arise because the memory was not written, was filtered by scope, was not retrieved by the vector index, was contradicted in the graph, or was selected but misused by the generator.

\paragraph{Consolidate and revise.}
Consolidation transforms raw traces into reusable abstractions,
\[
\text{trace} \rightarrow \text{summary} \rightarrow \text{fact} \rightarrow \text{relation}.
\]
Promotion should depend on repeated support, task utility, and risk:
\[
\operatorname{promote}(\mathcal{T}) =
\mathbb{I}\left[
\phi_{\mathrm{support}}(\mathcal{T})
+ \phi_{\mathrm{utility}}(\mathcal{T})
- \phi_{\mathrm{risk}}(\mathcal{T})
> \kappa
\right].
\]
Revision handles contradiction, decay, and correction:
\[
M_{t+1} = \operatorname{revise}(M_t, z_t).
\]
The system may increase confidence, reduce confidence, narrow scope, mark an object as superseded, or split an overly broad memory into more specific claims.
This makes memory plastic without treating every correction as destructive deletion.

\paragraph{Boundary projection.}
Before memory reaches the generator, MRMS performs boundary projection:
\[
\tilde{M}_t = \operatorname{project}(M_t; u,\pi_t,\mathcal{B}_t),
\]
where $\mathcal{B}_t$ denotes applicable boundary conditions such as subject, task mode, source class, privacy policy, or evidence channel.
Boundary projection is distinct from semantic retrieval.
A memory can be similar to the query and still be inappropriate for the current interaction.
Conversely, a memory can be weakly similar but necessary because it resolves a commitment or prevents a stale assumption from reappearing.
This motivates negative context: the context packet may include a compact statement that an older memory has been superseded or should not guide the current response.

\paragraph{Context packet.}
The output of selection is a structured packet rather than a flat prompt suffix.
MRMS separates local task state, selected episodic evidence, active semantic claims, unresolved uncertainty, external evidence, and negative context.
This packet design forces the generator to treat different memory classes differently.
For example, an episode can explain why a preference was inferred, while a semantic claim can state the current rule.
External evidence can ground factual claims without being mistaken for personal memory, and negative context can suppress a plausible but stale inference.

The packet is intentionally smaller than the retrieved candidate set.
It should contain enough evidence for the generator to act consistently, but not enough history to distract from the current task.
The selection operator therefore performs a compression step after graph expansion.
Repeated episodes can be summarized as a support count or short rationale.
Contradictions can be represented as a single uncertainty statement.
External evidence can be quoted or paraphrased with source labels, while personal memory remains separately marked.
This keeps the prompt budget aligned with the role of memory: memory should guide generation, not become an uncontrolled transcript replay.
Appendix~\ref{app:benchmark-details} lists the packet slots used by the prototype.

\paragraph{Substrate properties.}

The structured-vector-graph formulation gives MRMS several useful properties.
\textbf{Inspectability} follows from the structured record: every memory object has a textual claim, source class, scope, status, and revision history.
\textbf{Recall flexibility} follows from the vector index: semantically similar situations can be retrieved even when wording differs.
\textbf{Relational validity} follows from the graph: retrieved objects can be interpreted through support, contradiction, and supersession edges.
These properties are not redundant: removing the structured store makes memory difficult to govern, removing the vector store makes recall brittle, and removing the graph makes corrected memories difficult to distinguish from obsolete but semantically similar ones.
MRMS therefore treats retrieval as a controlled join across views and separates \emph{storage} from \emph{influence}: an object can be retained as evidence, embedded for future search, or linked as historical support while still being ineligible to guide current generation.
This separation matters because the cost of memory errors grows over time unless stale or overgeneralized memories are explicitly prevented from recurring.

\section{Evaluation Protocol}

A memory substrate should be evaluated by more than task completion.
We use six criteria: \textbf{continuity}, reuse of stable preferences and prior decisions; \textbf{specificity}, retrieval tied to the present input; \textbf{parsimony}, compact context exposure; \textbf{revision}, correction of outdated memory; \textbf{non-interference}, suppression of out-of-boundary memory; and \textbf{attribution}, source-aware use of context.

We evaluate substrate-level behavior with controlled scenarios that expose failures before generation: stale leakage, boundary contamination, unresolved contradiction, missing attribution, and inappropriate recall.
Ablations test whether each substrate mechanism is necessary under the constructed conditions.

Existing long-term memory benchmarks are useful starting points \citep{maharana2024locomo,wu2025longmemeval,hu2025memoryagentbench}, but our protocol isolates negative retrieval, revision, contradiction, and attribution.
It contains 800 deterministically generated synthetic tasks across delayed recall, boundary control, source separation, revision, stale suppression, evidence attribution, contradiction handling, and temporal commitment.
The prototype uses an in-memory structured store, a deterministic lexical-semantic vector index, a temporal policy, and a graph resolver, with no private data, production data, external API, or LLM judge.
Gold labels are memory identifiers, so the benchmark measures whether the substrate exposes the right evidence before generation.
The protocol asks whether structured gates prevent boundary contamination, temporal policies suppress stale or provisional memory, graph resolution recovers support and supersession evidence, and the resolver abstains when no scoped memory should influence generation; Appendix~\ref{app:benchmark-details} reports the full ablation.

\begin{center}
  \begin{minipage}{0.98\linewidth}
  \centering
  \small
  \setlength{\tabcolsep}{4pt}
  \begin{tabular}{lrrrrrr}
    \toprule
    System & Overall & Rev. & Stale & Bound. & Attr. & Abst. \\
    \midrule
    Recent context & 3.9 & 0.0 & 66.7 & 10.7 & 0.0 & 1.0 \\
    Vector only & 21.2 & 1.0 & 0.3 & 65.3 & 48.0 & 0.0 \\
    Structured + vector & 37.6 & 1.0 & 0.3 & 100.0 & 50.0 & 0.0 \\
    + temporal & 75.0 & 100.0 & 100.0 & 100.0 & 50.0 & 100.0 \\
    Full MRMS & 98.8 & 100.0 & 100.0 & 100.0 & 95.0 & 100.0 \\
    \bottomrule
  \end{tabular}
  \refstepcounter{table}
  \label{tab:results}
  \vspace{0.25em}

  \begin{minipage}{0.92\linewidth}
    \small
    \textbf{Table~\thetable:} Diagnostic benchmark over 800 tasks. Full MRMS obtains 98.8\% overall accuracy with a 95\% bootstrap interval of 98.0--99.4.
  \end{minipage}
  \end{minipage}
\end{center}

Table~\ref{tab:results} supports the narrow mechanism claim.
Vector retrieval improves recall but fails when the nearest memory is stale, out of scope, or unsupported.
Structured filtering improves boundary control; temporal policy repairs stale-memory suppression and abstention; graph expansion contributes most directly to contradiction handling and evidence attribution.
Full MRMS also keeps projection compact, selecting 1.11 memory objects on average after scoring 1.8 scoped candidates.

\paragraph{Qualitative analysis.}
Representative failures follow the ablation design.
Vector-only retrieval favored semantically similar but superseded memories; structured filtering removed unauthorized records but left contradictions unresolved.
Graph expansion recovered support and supersession evidence, with residual full-MRMS errors confined to evidence-attribution cases where the resolver selected the active claim but missed a designated support trace in 10 of 100 cases.

\section{Discussion and Future Work}

MRMS is a substrate-level framework: its contribution is the synchronization contract between structured records, vector retrieval, graph relations, temporal commitment, and boundary-aware context projection.
The prototype focuses on the pre-generation control points that determine what an LLM is allowed to use: selection, revision, boundary enforcement, and attribution.
This framing keeps the contribution independent of any single backend while making downstream agent integration testable.
Future evaluations should therefore report both downstream answers and the memory packet that preceded them, including rejected candidates, supersession traces, and attribution edges.
This makes MRMS suitable for studying not only whether an agent answers correctly, but why particular memories were permitted to shape the answer.
It turns future comparisons into an inspectable systems question rather than a black-box preference test.
Integrating the substrate with full LLM agents, learned retrievers, and real-world conversational benchmarks is the natural next stage of this research program.
In this view, memory becomes an auditable substrate for long-lived agency rather than a hidden prompt extension.

\bibliographystyle{plainnat}
\bibliography{references}

\appendix
\section{Additional Experimental Notes}
\label{app:benchmark-details}

\begin{table}[ht]
  \centering
  \small
  \begin{tabularx}{\linewidth}{@{}p{0.18\linewidth}p{0.25\linewidth}p{0.25\linewidth}X@{}}
    \toprule
    Layer & Structured form & Vector role & Graph role \\
    \midrule
    Short-term & Recent trace, active task state & Local semantic lookup & Links to current turn and unresolved references. \\
    Medium-term & Session summary, topic state & Retrieval across related sessions & Links decisions, open tasks, and derived summaries. \\
    Long-term & Stable claim, preference, constraint & Durable semantic recall & Tracks support, contradiction, supersession, and provenance. \\
    External evidence & Source-bounded document record & Evidence retrieval & Links claims to sources without absorbing them into personal memory. \\
    \bottomrule
  \end{tabularx}
  \caption{The temporal axis instantiated across the three representational views.}
  \label{tab:temporal-views}
\end{table}

\begin{table}[ht]
  \centering
  \small
  \begin{tabularx}{\linewidth}{@{}p{0.24\linewidth}p{0.34\linewidth}X@{}}
    \toprule
    Condition & Memory access & Competency tested \\
    \midrule
    Recent context only & Current prompt window & Baseline continuity without external memory. \\
    Vector retrieval & Nearest memory descriptors & Semantic recall and near-miss failures. \\
    Structured + vector & Source and scope gates before vector search & Boundary filtering and source separation. \\
    Structured + vector + temporal & Add status, layer, and recency policy & Revision, stale suppression, and abstention. \\
    Full MRMS & Add graph resolution and evidence expansion & Contradiction handling, supersession, and attribution. \\
    \bottomrule
  \end{tabularx}
  \caption{A compact ablation protocol for long-lived memory substrates.}
  \label{tab:protocol}
\end{table}

\begin{table}[ht]
  \centering
  \small
  \begin{tabularx}{\linewidth}{@{}p{0.24\linewidth}X@{}}
    \toprule
    Packet slot & Contents \\
    \midrule
    Local state & Current task, recent turns, unresolved references, and active constraints. \\
    Episodic evidence & Time-stamped traces that justify why a memory may matter now. \\
    Semantic claims & Active durable preferences, facts, commitments, or constraints. \\
    Relation notes & Support, contradiction, supersession, and scope links from the memory graph. \\
    External evidence & Source-bounded retrieved material separated from personal memory. \\
    Negative context & Compact statements about rejected, stale, or superseded assumptions. \\
    \bottomrule
  \end{tabularx}
  \caption{Context packet slots exposed to the generator after selection.}
  \label{tab:context-packet}
\end{table}

\paragraph{Data generation.}
The diagnostic benchmark contains 800 generated tasks with 100 tasks per family.
Each scenario contains a query, a set of memory records, typed metadata, optional graph edges, and gold memory identifiers.
The generator samples subjects, task modes, values, layers, timestamps, sources, statuses, and distractor records from fixed vocabularies under seed 20260702.
The average scenario contains 17.0 memory records, with a minimum of 13 and a maximum of 21.
Across the generated benchmark, 9,503 records are personal memories and 4,095 are external guideline records.
The benchmark uses synthetic text only and includes no private data, production traces, external API calls, or LLM-generated judgments.
The reproducibility artifacts consist of scenario records, decision traces, per-task metrics, and aggregate summaries.

\paragraph{Task families.}
Delayed-recall tasks test whether a system can retrieve a durable scoped memory after irrelevant distractors are added.
Near-miss boundary tasks add semantically similar memories attached to the wrong subject or task.
Source-boundary tasks distinguish personal memories from external guidelines.
Revision tasks include an old memory and a newer superseding memory.
Stale-suppression tasks require abstention when no active memory remains.
Evidence-attribution tasks require selecting both a durable claim and at least one supporting trace.
Contradiction tasks include conflicting memories with typed support and contradiction edges.
Temporal-commitment tasks distinguish recent provisional traces from older durable commitments.

\paragraph{Ablations.}
The recent-context baseline can inspect only active short-term traces.
The vector-only baseline performs user-scoped nearest-neighbor retrieval but ignores source, task scope, status, temporal layer, and graph relations.
The structured-vector condition applies exact gates over user, subject, task, and source before vector search.
The structured-vector-temporal condition adds active-status filtering, layer-aware ranking, stale suppression, and abstention.
The full MRMS resolver additionally traverses \emph{supersedes}, \emph{contradicts}, \emph{supports}, and \emph{derived-from} edges before assembling the context packet.

\paragraph{Metrics.}
Overall accuracy requires the pre-generation selected memory set to include the gold memory identifiers and exclude stale or forbidden identifiers.
Revision success measures whether superseded memories are suppressed in favor of active replacements.
Stale suppression measures whether inactive memory is not exposed when no active scoped memory exists.
Boundary control measures whether source, user, subject, and task gates prevent near-miss contamination.
Attribution measures whether evidence-bearing tasks expose a supporting trace rather than only the final claim.
Abstention accuracy measures whether the system returns no memory when no valid scoped memory should influence generation.
Auxiliary statistics track selected memory count, a synthetic context-token proxy, and the number of candidates scored after structured gates.
Bootstrap confidence intervals are computed over generated tasks.

\begin{table}[ht]
  \centering
  \small
  \setlength{\tabcolsep}{3.5pt}
  \begin{tabular}{lrrrrr}
    \toprule
    Family & Recent & Vector & S+V & +Temp. & Full \\
    \midrule
    Delayed recall & 0 & 73 & 100 & 100 & 100 \\
    Near-miss boundary & 30 & 0 & 100 & 100 & 100 \\
    Source boundary & 0 & 96 & 100 & 100 & 100 \\
    Revision & 0 & 1 & 1 & 100 & 100 \\
    Stale suppression & 1 & 0 & 0 & 100 & 100 \\
    Evidence attribution & 0 & 0 & 0 & 0 & 90 \\
    Contradiction & 0 & 0 & 0 & 0 & 100 \\
    Temporal commitment & 0 & 0 & 0 & 100 & 100 \\
    \bottomrule
  \end{tabular}
  \caption{Per-family accuracy by ablation condition. Values are percentages over 100 tasks per family. S+V denotes structured-vector retrieval.}
  \label{tab:per-family-results}
\end{table}

\paragraph{Interpretation.}
The appendix documents how the diagnostic result was produced and why the ablations isolate different substrate components.
The benchmark is organized around pre-generation memory decisions because this is where MRMS differs from ordinary retrieval pipelines.
Each run can be inspected as a decision trace: eligible records after structured gates, ranked vector candidates, graph expansions, rejected records, and final context-packet slots.
These traces make the measured outcomes auditable.
A stale-suppression success can be traced to status authority, an attribution success to support edges, and a boundary-control success to scope and source gates.
Future evaluations can attach the same trace format to LLM-generated answers, allowing downstream quality to be analyzed together with the memory decisions that shaped it.
Natural extensions include replacing the lexical vector proxy with learned embeddings, adding noisier multi-turn conversations, and measuring generated answers with human or task-specific judgments.
The same artifacts also support error analysis by separating missing gates, incorrect ranking, absent graph edges, and packet-assembly mistakes.
This preserves comparability across future implementations while keeping the diagnostic target tied to the substrate itself.

\end{document}